\pdfoutput=1
\documentclass{article}

\usepackage[final]{neurips_2020}
\usepackage{textcomp}
\usepackage[usenames,dvipsnames,dvinames]{xcolor}
\usepackage{url,times,stfloats,flushend,float,crop,color,chngpage,array,alltt}
\usepackage{amsmath,amsthm,amsfonts,graphicx,amssymb,sidecap,color,wrapfig,microtype,cancel,
  bbm, bm}
\usepackage{caption, subcaption}
\usepackage{algorithm}
\usepackage{algpseudocode}
\usepackage{multicol}
\usepackage{multirow}
\usepackage{enumitem}
\usepackage{eqparbox}

\newcommand\LONGCOMMENT[1]{%
  \hfill//\ \begin{minipage}[t]{\eqboxwidth{COMMENT}}\emph{#1}\strut\end{minipage}%
}
\newcommand{\algred}[1]{{\color{red} #1}}
\DeclareMathOperator*{\argmin}{argmin}
\newcommand{\RR}{\mathbb{I\!\!R}} 
\newcommand{\ba}{\bm{a}}
\newcommand{\bz}{\bm{z}}
\newcommand{\balpha}{\bm{\alpha}}
\newcommand{\bzhat}{\hat{\bm{z}}}

\newcommand{\bd}{\bm{d}}
\newcommand{\bx}{\bm{x}}
\newcommand{\by}{\bm{y}}
\newcommand{\bw}{\bm{w}}
\newcommand{\bv}{\bm{v}}
\newcommand{\Id}{\mathcal{I}}
\newcommand{\indicator}{\mathbf{1}}
\newcommand{\librsb}{$\mbox{{\sf librsb}}$}
\newcommand{\pytorch}{$\mbox{{\sf PyTorch}}$}
\newcommand{\tensorflow}{$\mbox{{\sf TensorFlow}}$}
\newcommand{\liblinear}{$\mbox{{\sf LIBLINEAR}}$}

\newcommand{\sklearn}{$\mbox{{\sf scikit-learn}}$}

\newcommand{\omp}{$\mbox{{\sf OpenMP}}$}
\newcommand{\cuda}{$\mbox{{\sf CUDA}}$}
\newcommand{\thrust}{$\mbox{{\sf Thrust}}$}
\newcommand{\cublas}{$\mbox{{\sf cuBLAS}}$}
\newcommand{\cusparse}{$\mbox{{\sf cuSPARSE}}$}
\newcommand{\tronlr}{\emph{TRON}-LR}
\newcommand{\tronminus}{\emph{TRON}-LR-GPU$^0$}
\newcommand{\tronlrg}{\emph{TRON}-LR-GPU}
\newcommand{\tronlrc}{\emph{TRON}-LR-CPU}
\newcommand{\tronlrm}{\emph{TRON}-LR-MIX}
\newcommand{\lrL}{f_{\mbox{LR}}}
\newcommand{\svmL}{f_{\mbox{L2}}}
\newcommand{\reg}{$L_2$}
\newcommand{\lsvm}{L2-SVM}
\newcommand{\loss}{\ell}
\newcommand{\tron}{\emph{TRON}}
\newcommand{\tronsvm}{\emph{TRON}-SVM}
\newcommand{\tronsvmg}{\emph{TRON}-SVM-GPU}
\newcommand{\tronsvmc}{\emph{TRON}-SVM-CPU}
\newcommand{\tronsvmm}{\emph{TRON}-SVM-MIX}
\newcommand{\cgls}{\emph{L2-SVM-MFN}}
\newcommand{\lbfgs}{\emph{L-BFGS}}
\newcommand{\lrD}{D}
\newcommand{\svmD}{D}
\newcommand{\twonorm}[1]{\ensuremath{\frac{1}{2}\|#1\|^2_2}}

\title{GPU-Accelerated Primal Learning for Extremely Fast
Large-Scale Classification}

\author{{\bf John T. Halloran} \\
Department of Public Health Sciences \\
University of California, Davis \\
\texttt{jthalloran@ucdavis.edu} \\
\And
{\bf David M. Rocke} \\
Department of Public Health Sciences \\
University of California, Davis \\
\texttt{dmrocke@ucdavis.edu}
}

\usepackage{hyperref}
\hypersetup{bookmarks,colorlinks=true,citecolor=Violet,linkcolor=Mahogany,urlcolor=blue}
\begin{document}

\maketitle

\begin{abstract}
One of the most efficient methods to solve \reg{}-regularized
primal problems, such as logistic regression and linear support vector
machine (SVM) classification, is the widely used trust region Newton
algorithm, \tron{}~\cite{lin2008trust}.
While \tron{} has recently been shown to enjoy substantial speedups on shared-memory multi-core
systems~\cite{lee2015fast, halloran2018matter}, exploiting
graphical processing units (GPUs) to speed up the method
is significantly more difficult, owing to the highly complex
and heavily sequential nature of the algorithm.  
In this work, we show that using judicious GPU-optimization
principles, \tron{} training time for different losses and
feature representations may be drastically reduced.  For sparse feature
sets, we show that using GPUs to train
logistic regression classifiers in \liblinear{} is up to an
order-of-magnitude faster than solely using multithreading.  For dense
feature sets--which impose far more stringent memory constraints--we show
that GPUs substantially reduce the lengthy SVM learning times required for
state-of-the-art proteomics analysis,
leading to
dramatic improvements over recently proposed speedups.  Furthermore,
we show how GPU speedups may be mixed with multithreading to enable
such speedups when the dataset is too large for GPU memory requirements; on
a massive dense proteomics dataset of nearly a quarter-billion data
instances, these mixed-architecture speedups
reduce SVM analysis time from over half a week to less than a
single day while using limited GPU memory.  

\end{abstract}

\section{Introduction}\label{section:intro}
Over the past decade, GPUs have become 
valuable computing resources to accelerate the training of
popular machine learning models, playing a key role in the widespread use
of deep models and the growing ecosystem of deep learning
packages~\cite{chen2015mxnet, chollet2015keras,
  tensorflow2015-whitepaper, jia2014caffe, pytorchNeurips2019}.  When a training algorithm
admits an efficient GPU implementation (such as gradient
boosted trees~\cite{mitchell2017accelerating}, nonlinear kernel
learning~\cite{catanzaro2008fast, wen2018thundersvm}, and primal
methods like \lbfgs{}~\cite{liu1989limited} and variants of gradient
descent), the speedups gained using GPUs, as opposed to only CPUs, are
often substantial.  For instance, in \pytorch{}~\cite{pytorchNeurips2019},
training a logistic regression classifier on the
\texttt{rcv1}~\cite{lewis2004rcv1}
dataset with gradient descent is $14.6$ times faster using a
Tesla V100 GPU versus using 24 CPU threads with an Intel Xeon Gold
5118 (similarly, training with \lbfgs{} in this example is $13.1$
times faster using the V100, detailed in Appendix~\ref{appendix:pytorch}).

Specialized solvers commonly provide even more speed.  For the
previous logistic regression example, using just a single CPU thread
with \sklearn{}'s~\cite{scikit-learn} \tron{} solver--the primal
learning algorithm for logistic regression and SVM
classification/regression adapted from
\liblinear{}~\cite{fan2008liblinear}--is $94.7$ and $10.9$ times
faster than GPU-accelerated gradient descent and \lbfgs{},
respectively, implemented in \pytorch{}. 
However, while significant work has been done to further
accelerate \tron{} and many other extremely fast
machine learning solvers~\cite{hsieh2008dual, 
  yuan2012improved, keerthi2005modified, johnson2015blitz,
  johnson2018fast} using multiple CPU
cores~\cite{bradley2011parallel, jaggi2014communication,
  hsieh2015passcode, lee2015fast, chiang2016parallel, smith2017cocoa,
  zhuang2018naive, halloran2018matter},
analogous GPU speedups for such efficient algorithms are typically
lacking.  This lack of GPU exploitation is due to the specialized
structure and complexity of these algorithms, which
{\bf naturally lend themselves to multithreaded speedups} on shared memory
systems, {\bf yet resist optimizations on GPU architectures}.

For example, \tron{} relies on random access to features for SVM
losses, which is naturally supported in
multithreaded systems, but prevents memory coalescing
(and is thus deleterious) for GPU
computation.  Furthermore, large memory transfers between the CPU and
GPU are expensive, so that the complex, sequential dependency of
variables in specialized algorithms like \tron{} make optimal GPU use
difficult.  Indeed, we show that while most of the computational
bottlenecks for logistic regression in \tron{} are linear algebra
operations~\cite{lee2015fast} (for which GPUs greatly outperform
CPUs), using \tron{}'s original variable access pattern in
\liblinear{} results in poor GPU performance--performing even worse
than using only a single CPU thread on one of the
presented datasets.

Herein, we show that using just a
single GPU, excellent training speedups are achievable by overly CPU-specialized
machine learning algorithms such as \tron{}.  In particular, for
different feature representations and loss functions, we show that
\tron{} training times may be drastically reduced using judicious
GPU-optimization principles.  

{\bf Sparse Features.}  For sparse feature representations, we
successively optimize \tron{} for logistic regression (referred to as \tronlr{}) in \liblinear{} using several strategies to:
a) decouple the sequential dependence of variables, b)
minimize the number of large-memory transfers between GPU and CPU,
and c) maximize parallelism between the CPU and GPU.  We show that
while \tron{}'s original variable access pattern limits the
effectiveness of GPU computation, using a single CPU thread with our
GPU optimizations results in a  70.8\% improvement in training time
(averaged over the presented datasets) over the single-threaded
version of \tron{} in standard \liblinear{}.  In addition, we show
that mixing our GPU optimizations with multithreading provides further speedups,
resulting in an average {\bf 89.2\% improvement over single-thread optimized \tron{}} and an
average {\bf 65.2\% improvement over \tron{} in the
multithread-optimized version of \liblinear{}}~\cite{lee2015fast}.

{\bf Dense Features.}  For dense feature representations, we show that GPUs 
substantially reduce SVM learning times for
state-of-the-art analysis of dense proteomics
datasets~\cite{kallPercolator}. Overcoming the random access
restrictions of \tron{} SVM learning (referred to as \tronsvm{}), we
show that using just a single GPU leads to an average 
as much as triples the 
performance of recently proposed speedups for this
application~\cite{halloran2018matter}.  {\bf On a large-scale dataset
  of over 23 million data instances, these GPU speedups reduce SVM
  learning time from 14.4 hours down to just 1.9 hours}. Furthermore,
dense feature sets impose stringent GPU memory constraints,
particularly for the massive datasets regularly produced in
biological experiments.  Thus, we demonstrate how GPU optimizations may
be mixed with multithreading to significantly reduce GPU memory
constraints.  On a massive proteomics dataset consisting of over
215 million data instances--which exceeds memory requirements for
GPU-only speedups--these mixed-architecture speedups drastically
outperform recent multithread-optimized solvers, {\bf reducing
standard analysis time from 4.4 days down to just 19.7 hours}.

The paper is organized as follows.  In
Section~\ref{section:previousWork}, we describe relevant previous work
speeding up \tron{} for both sparse and dense feature sets on shared
memory systems.  In Section~\ref{section:tron}, we define the
general \tron{} algorithm and computational bottlenecks encountered
minimizing different loss functions.  In
Sections~\ref{section:tronlr} and~\ref{section:svmGpus}, we discuss
how the computational bottlenecks in algorithms like \tron{} natively
resist GPU speedups, and GPU-optimization principles to
overcome these hurdles (providing the resulting GPU optimizations for
the objectives and feature architectures under study).  We demonstrate
that the presented GPU-optimizations drastically outperform recent
multithreaded speedups in Section~\ref{section:results}, and conclude
with future avenues extending the presented work to other
high-performance GPU packages (such as \pytorch{}) in
Section~\ref{section:conclusions}.



\vspace{-0.05in}
\section{Previous Work}\label{section:previousWork}
\vspace{-0.05in}
Serving as the primal solver in the popular package
\liblinear{}~\cite{fan2008liblinear}, \tron{} has been extensively
tested and shown to enjoy superior
speed and convergence compared to other second-order solvers, such as
the widely-used quasi-Newton algorithm \lbfgs{}~\cite{liu1989limited}
and the modified Newton algorithm
\cgls{}~\cite{keerthi2005modified} (one of the fastest 
algorithms for large-scale primal SVM learning).  As a Newton method,
the algorithm enjoys general quadratic convergence without loading the
entire Hessian into memory, thus only using linear memory.  For
logistic and SVM losses in \liblinear{}, \tron{}'s convergence speed
has further been theoretically
improved by refining trust-region update rules~\cite{hsiastudy}
and applying a preconditioner matrix to help stabilize
optimization~\cite{hsia2018preconditioned}.  In~\cite{lee2015fast},
multithreaded optimizations in shared-memory multi-core systems were
extensively explored to speed up \tron{}'s computational bottlenecks
(further described in Section~\ref{section:tron}) for logistic
regression.  Evaluating several multithreading libraries (i.e.,
\omp{}, Intel's Math Kernel Library, and the sparse matrix
multiplication package \librsb{}) over a large number of datasets,
\omp{} was found to provide the best multithreaded performance and was
subsequently integrated into the multi-core release of \liblinear{}.

\subsection{SVM Classification Using \tron{} for Fast Large-Scale
  Proteomics Analysis}\label{section:percolator}
In proteomic analysis pipelines, SVM classification using
Percolator~\cite{kallPercolator} is a critical step towards accurately
analyzing protein data collected via tandem mass spectrometry (MS/MS).  Given
a collection of MS/MS spectra representing the protein
subsequences (called \emph{peptides}) present in a biological
sample, the first stage of proteomics analysis typically consists of
identifying the input spectra by searching (i.e., scoring and ranking)
a database of peptides.  This first stage thus results in a list of
\emph{peptide-spectrum-matches} (\emph{PSMs}) and their respective
scores.  In practice, however, database-search scoring functions are
often poorly calibrated, making PSMs from different spectra difficult
to compare and diminishing overall identification accuracy.  
To correct for this, the list of PSMs, as well as dense feature
vectors describing each match, are fed into Percolator for
\emph{recalibration}.

Percolator first estimates PSM labels
(i.e., correct versus incorrect) using false discovery rate
analysis~\cite{kall2008assigning}, then trains a linear SVM to
classify correct and incorrect identifications.  These two steps are
iterated until convergence and the input PSM
scores are subsequently recalibrated using the final learned
SVM parameters.  Furthermore, to prevent overfitting and improve
generalizability within each iteration, three-fold cross-validation
is carried out over three disjoint partitions of the
original dataset, followed by further nested cross-validation within
each fold~\cite{granholm2012cross}.


The accuracy improvements of Percolator recalibration have been well
demonstrated for a wide variety of PSM scoring functions--e.g.,
linear~\cite{kallPercolator, broschAccurate, xu2013combining}, $p$-value
based~\cite{granholm2013fast, howbert:computing, lin2018combining},
and dynamic Bayesian networks~\cite{halloran2014uai-drip,
  halloran2016dynamic, halloran2018analyzing}--and complex
PSM feature sets--e.g., Fisher kernels~\cite{halloran2017gradients,
  halloran2018learning}, subscores of linear
functions~\cite{spivak:learning}, ensembles of scoring
functions~\cite{wen2015ipeak}, and features derived using deep
models~\cite{gessulat2019prosit}.  However, due to the iterative
training of many SVMs during cross-validation, Percolator requires
substantial analysis times for large-scale datasets commonly
collected in MS/MS experiments.  Initial work sought to
speed up Percolator runtimes by randomly sampling a small portion of
the data to train over~\cite{maccoss2016fast}, but this was
subsequently shown to unpredictably diminish the performance of learned
parameters~\cite{halloran2018matter}.  Thus, to combat these lengthy
analysis times without affecting learned SVM parameters, recent
work~\cite{halloran2018matter} applied extensive systems-level
speedups and multithreading in both Percolator's original primal
solver, \cgls{}, and \tron{} (heavily optimized to utilize dense
feature vectors).  While both optimized solvers were shown to
significantly improve Percolator training times for large-scale data,
\tron{} displayed markedly superior performance. 

\vspace{-0.05in}
\section{Trust Region Newton Methods for Primal
  Classification}\label{section:tron}
\vspace{-0.05in}
Consider feature vectors $\bx_i \in \RR^n, i = 1, \dots, l$ and label vector
$\by \in \{-1, 1\}^l$, and  let $X = [\bx_1 \dots 
  \bx_l]^T$ be the feature matrix.  For vectors, index-set subscripts
  denote subvectors and for matrices,
pairs of index-set subscripts denote submatrices.  The general
objective, which we wish to minimize w.r.t. $\bw$, is
\begin{align}
f(\bw) =& \twonorm{\bw} + C \sum_{i = 1}^l \loss(\bw;\bx_i, y_i),\label{eq:loss}
\end{align}
where $\twonorm{\bw}$ is the regularization term, $C > 0$ is a
regularization hyperparameter, and $\loss(\bw;\bx_i, y_i)$ is a loss
function.  

When $\loss(\bw;\bx_i, y_i) = \log ( 1 +
\exp{ (-y_i \bw^T \bx_i) } )$, commonly referred to as the \emph{logistic
loss}, minimizing Equation~\ref{eq:loss} corresponds to learning a
classifier using logistic regression.  Similarly, minimizing
Equation~\ref{eq:loss} when $\loss(\bw;\bx_i, y_i)= (\max(0,1 - y_i
\bw^T \bx_i))^2$, commonly referred to as the \emph{quadratic SVM} or \emph{\lsvm{}
loss}, corresponds to learning a linear SVM classifier.  
  We denote Equation~\ref{eq:loss}
under the logistic loss as  $\lrL (\bw)$ and, under the \lsvm{} loss, as
$\svmL (\bw)$.  

\begin{algorithm}
\caption{The \tron{} algorithm}\label{algorithm:tron}
\begin{algorithmic}[1]\small
\State Given $w$, $\Delta$, and $\sigma_0$
\State Calculate $f(\bw)$ \Comment{Critically depends on $\bz =
  X^T\bw$}
\While{Not converged}
\State Find $\bd =  \argmin_{\bv}q(\bv)\, \mbox{ s.t. }
\lVert \bv \rVert_2 \leq \Delta.$ \LONGCOMMENT{Critically depends on
  $\nabla f(\bw)$, $\nabla^2 f(\bw) \bv$}
\State Calculate $f(\bw + \bd), \sigma = \frac{f(\bw + \bd) -
  f(\bw)}{q(\bd)}$ \Comment{Critically depends on $\bz =
  X^T(\bw + \bd)$}
\If{$\sigma > \sigma_0$}
\State $\bw \leftarrow \bw + \bd$, increase trust region $\Delta$.
\Else \quad 
\State Shrink $\Delta$.
\EndIf
\EndWhile
\end{algorithmic}
\end{algorithm}

\tron{} is detailed in Algorithm~\ref{algorithm:tron}.  At each
iteration, given the current parameters $\bw$ and trust region
interval $\Delta$, TRON considers the following quadratic
approximation between function parameters,
\begin{align}
f(\bw + \bd) - f(\bw) \approx q(\bd) \equiv \nabla
f(\bw)^T\bd + \frac{1}{2}\bd^T \nabla^2
f(\bw)\bd.\label{eq:quadraticApprox}
\end{align}
A truncated Newton step ($\bd$ on line 4 in
Algorithm~\ref{algorithm:tron}), confined in the trust region, is then
found using a conjugate gradient procedure.  
If $q(\bd)$ is close to $f(\bw + \bd)
- f(\bw)$, $\bw$ is updated to $\bw + \bd$ and the trust region
interval is increased for the subsequent iteration.  Otherwise, $\bw$
remains unchanged and the trust region interval is shrunk.  

Note that the function evaluation $f(\bw)$--which critically depends
on computing $\bz = X^Tw$ for both losses-- must be computed for each new
iteration, as well as the gradient and Hessian for
Equation~\ref{eq:quadraticApprox}.  However, computing only the
Hessian-vector product in 
Equation~\ref{eq:quadraticApprox} avoids loading
the entire Hessian into memory (which would be intractable for large
datasets).  Thus, the most intensive portions of \tron{} are the
computations of $\bz = X^Tw, \nabla f(\bw),$ and $\nabla^2 f(\bw) \bv$
(where $\bv$ is the optimization variable in line 4 of
Algorithm~\ref{algorithm:tron}),
summarized for both losses in Table~\ref{table:bottleneckOperations}.
Further derivation of these quantities is available in Appendix~\ref{section:derivation}.

We note that arbitrary loss functions (and combinations
thereof) may be used in Equation~\ref{eq:loss}, thus allowing future
 work utilizing the highly efficient \tron{} in
popular automatic differentiation~\cite{baydin2017automatic}
packages~\cite{pytorchNeurips2019, tensorflow2015-whitepaper,
  tokui2015chainer, neubig2017dynet}.  However, these packages rely on
GPUs for optimal performance, the use of which \tron{}
natively resists (as we'll see, and rectify, for the two loss
functions considered).



\begin{table}[htbp!]
\small
\centering
\setlength\tabcolsep{2pt}
\renewcommand{\arraystretch}{2}
\begin{tabular}{p{0.5\textwidth}|p{0.5\textwidth}}
\multicolumn{1}{c}{\bf Logistic Loss} & \multicolumn{1}{c}{\bf
  \lsvm{} Loss}\\\hline
\multicolumn{1}{c}{$\bz = X^T w$, to compute $\lrL (\bw)$}
& 
\multicolumn{1}{|c}{$\bz = X^T w$, to compute $\svmL (\bw)$}\\\hline
$\nabla
\lrL (\bw) = \bw + C
\sum_{i=1}^l (h(y_i \bz_i) - 1) y_i \bx_i$, where $h(y_i
\bz_i) = (1 + e^{-y_i \bz_i})^{-1}$ & $\nabla \svmL (\bw) =
\bw +
2CX_{I, :}^T(\bz_{I} - \by_{I})$, where $I \equiv \{ i | 1 - y_i
\bz_i > 0 \}$ is an index set and the operator $:$ denotes
all elements along the corresponding dimension (i.e., all columns in
this case)\\\hline
$\nabla^2 \lrL (\bw) \bv = \bv + C
X^T(\lrD (X\bv))$, where $\lrD$ is a diagonal matrix with elements
$\lrD_{i,i} = h(y_i \bz_i)(1 - h(y_i \bz_i))$ & $\nabla^2 \svmL (\bw)
\bv = \bv + 2 C
X_{I,:}^T(X_{I,:}\bv)$\\\hline
\end{tabular}
\caption{\tron{} major bottleneck computations for logistic and \lsvm{} losses.}
\label{table:bottleneckOperations}
\end{table}
\vspace{-0.1in}
\section{Accelerating \tronlr{} training using GPUs}\label{section:tronlr}
\vspace{-0.1in}
Assume a shared-memory
multi-core system and a single GPU with sufficient memory for the
variables in Table~\ref{table:bottleneckOperations} (this is later
relaxed in Section~\ref{section:svmGpus}).  Herein, the CPU is
referred to as the \emph{host} and the GPU is referred to as the \emph{device}.

\tronlr{} runtime is dominated by three major matrix-vector
multiplications in the bottleneck computations listed in
Table~\ref{table:bottleneckOperations}: $\bz = X^T w, \nabla^2 \lrL (\bw)
\bv$, and $\nabla \lrL (\bw) = \bw  + C X\bzhat$, where $\bzhat_i =
(h(y_i \bz_i) - 1) y_i$.  For instance, profiling \tronlr{} in
\liblinear{} training on the large-scale
\texttt{SUSY}~\cite{baldi2014searching} dataset, these {\bf three
matrix-vector multiplications account for 82.3\% of total training
time}.  We thus first attempt to accelerate \tronlr{} by computing
these quantities quickly on the device (as was similarly done
in~\cite{lee2015fast} using multithreading).

In \liblinear{}, this first attempt
at GPU acceleration (called \tronminus{}) is implemented using
\cusparse{} to perform sparse linear algebra operations as
efficiently as possible for \liblinear{}'s sparse feature representation.  Compared to
the standard single-threaded implementation of \liblinear{} on the
\texttt{SUSY} dataset, \tronminus{} achieves a speedup of
0.65--\tronminus{} is actually slower than the single-threaded
\liblinear{}!
\tronminus{} fairs better on other presented datasets, but
performs poorly overall (displayed in Figure~\ref{fig:lrTiming}).

\subsection{Sequentially Dependent Variables}\label{section:dependencies}
The critical issue encountered by \tronminus{} is \tron{}'s overly
sequential dependency of variables; once variable vectors are
computed on the GPU, they are immediately needed on the host CPU to
proceed with the next step of the algorithm.  For instance, computing the bottleneck
$z = X^T(\bw+\bd)$ using \cusparse{} is fast, but $z$ must immediately be
transferred back to the host to compute $\lrL(\bw+\bd)$ (in line 5 of
Algorithm~\ref{algorithm:tron}).  However, large-memory transfers
between the host and device are expensive, especially when either the
host or device are waiting idle for the transaction to complete
(steps to conceal transfer latency are discussed
in Appendix~\ref{section:concealing}.  




Furthermore, all other major operations in
Algorithm~\ref{algorithm:tron} are locked in the same manner as the
previous bottleneck example: the trust region update (lines 6-10) can
not proceed without the value of $\lrL(\bw+\bd)$, and, without either
the updated $\bw$ or trust region, operations for the next iteration's
truncated Newton step (line 4) are unable to run concurrently in an
attempt to conceal transfer latency.  Clearly, this pattern of
variable access is suboptimal for GPU use (best evidenced by
\tronminus{}'s performance in Section~\ref{section:results}).



\subsection{Decoupling Dependencies to Maximize Host and Device
  Parallelism}\label{section:gpuOpt}
To optimally use the GPU, we must first decouple the sequential
dependency of variables discussed in
Section~\ref{section:dependencies}.
Recall that, for $\nabla \lrL (\bw)$, the vector $\bzhat$ is such that
$\bzhat_i = (h(y_i \bz_i) - 1) y_i$.  To decrease sequential
dependencies on the computational bottleneck
$z = X^T(\bw+\bd)$, we first note that
calculation of $\lrL(\bw+\bd)$ always precedes $\nabla
\lrL(\bw+\bd)$.  Thus, to decouple gradient variables, once $z$ is
calculated on the device, we prepare all device-side variables needed
to compute $X^T\bzhat$ in the event that
$\sigma > \sigma_0$. 
Specifically,  after computing $z = X^T(\bw+\bd)$ on the device, we
use a custom \cuda{} kernel to compute $\bzhat$ followed by a \thrust{}
reduction to compute 
$\lrL(\bw+\bd) = \twonorm{\bw+\bd} + C \sum_{i   =
  1}^l \log ( 1 + e^{ -y_i \bz_i })$ (note that the scalar output of the
reduction, i.e. $\lrL(\bw+\bd)$, is immediately available to the
host).  The computation of $\bzhat$ is massively parallelizable, so
the grid-stride loop in the custom kernel is extremely efficient.  Thus, if
$\sigma > \sigma_0$, the variable  $\bzhat$ is already in
device memory and the gradient is quickly calculated using \cusparse{}
on the device as $\nabla \lrL(\bw+\bd) = \bw+\bd + X^T \bzhat$.  Finally,
$\nabla \lrL(\bw+\bd)$ is transferred from device to host, which is
notably efficient when $l \gg n$ (i.e., the optimal setting for
primal learning).

This set of operations accomplishes several optimizations
simultaneously:
\begin{itemize}[leftmargin=20pt,itemsep=-0.1ex] 
\item {\bf Decoupling dependencies, avoiding large transfers}:
$\bz$ and $\hat{\bz}$ are completely decoupled of any dependency for
 host-side computation, thanks to the custom reduction and kernel.
 This saves several large transfers of $\bz, \hat{\bz}$ from (and to)
 the device, and avoids the need to conceal transfer latency.
\item {\bf Coalesced memory}: the device performs optimally as all
  operations allow memory coalescing.
\item {\bf Device saturation}: an uninterrupted series of intensive
  computation is performed on the device (i.e., no device-side stalls due
  to host dependencies).
\item {\bf Host and device parallelism}: the complete decoupling of
  $\bz, \hat{\bz}$ allows more independent operations to be run on the host while
  the device runs concurrently.
\end{itemize}

We complete the total GPU optimization of \tronlr{} by speeding up the
remaining bottleneck, the Hessian-vector product $\nabla^2 \lrL (\bw) \bv$.
As with the previous optimizations, device variables are maximally decoupled
from host-side dependencies, while using device-side functions which
allow peak performance.  Further details are
in Appendix~\ref{section:hessianVectorProducts}, and a comprehensive summary of the
previously described \tronlr{} GPU-optimizations is listed in Appendix~\ref{section:summary}.

{\bf Decreasing runtimes via mixed-architecture
  speedups.}
While computations remain which
may be accelerated using the same GPU-optimization principles,
allocating additional device vectors becomes
problematic for large-scale datasets and current GPU memory
ranges.  Thus, in addition to the
previously described GPU optimizations, we accelerate remaining
bottleneck areas using multithreading.  In particular, multithreading
using \omp{} is used to accelerate vector-matrix-vector
multiplications in the conjugate gradient procedure (previously
optimized using loop unrolling) and application of the preconditioner
matrix~\cite{hsia2018preconditioned} (which is jointly sped up using
existing device-side computation during the Hessian-vector
product optimizations).
\vspace{-0.1in}
\section{Accelerating \tronsvm{} training using
  GPUs}\label{section:svmGpus}
\vspace{-0.1in}
\begin{table}
\small
\centering
\setlength\tabcolsep{2pt}
\begin{tabular}{p{0.5\textwidth}|p{0.5\textwidth}}
\multicolumn{1}{c}{{\bf \tronsvmg}} & \multicolumn{1}{c}{\bf \tronsvmm}\\\hline
$\bz =X\bw$  is calculated and stored on the device.& $\bz =X\bw$
is calculated and stored on the device, then transferred to the
 host.\\\hline
$I =
\{i: y_i\bz_i < 1\}$ is calculated on the device, then $\svmL (\bw) =
\twonorm{\bw} + C \sum_{i = 1}^l(1 - y_i\bz_i > 0)^2$ is computed on
the device while the host runs independent, sequential operations. &
$I$ is calculated on the device, then transferred to the
host.  The device-side computation of
$\svmL (\bw)$ is run concurrently with this transfer.\\\hline
On the device, $\hat{\bz} = (\bz_I - \by_I)$ and
  $\hat{X} = X_{I,:}$ are computed.
  The gradient $\nabla \svmL (\bw) = \bw +  2C\hat{X}^T\hat{\bz}$ is
  then computed and transferred to the host.
&
With $I$ and $z$ on the host, $\nabla \svmL (\bw)$ is computed using
multithreading.
\\\hline
The Hessian-product is computed on the device as $\nabla^2
  \svmL (\bw) \bv = \bv + 2 C \hat{X}^T(\hat{X}\bv)$ and
  transferred to the host.
& Using multithreading, the Hessian-product is calculated on the
  host as $\nabla^2 \svmL (\bw) \bv = \bv + 2 C
  X_{I,:}^T(X_{I,:}\bv) = \bv + 2C\sum_{i \in
    I}(\bx_i^T\bv)\bx_i$.
\\\hline
\end{tabular}
\caption{Major operations of the \tronsvm{} solvers designed for GPU
  compute.}
\label{table:tronGpuSolvers}
\end{table}

Focusing on speeding up SVM learning in the state-of-the-art software
Percolator~\cite{kallPercolator}--which uses dense feature vectors to
analyze large-scale proteomics datasets--the GPU-optimization
principles from Section~\ref{section:gpuOpt} are applied to
\tronsvm{}: device-side variables are decoupled from dependent
host-side computations, necessary transfers are run asynchronously in
parallel with the maximum number of host/device operations,
and linear algebra operations are extensively and efficiently carried
out using \cublas{}.  However, speed ups in \tronsvm{} possess a key
difficulty for GPU computation; for $\bz = X^T w$, the
active set $I \equiv \{ i | 1 - y_i \bz_i > 0 \}$ is recomputed every
iteration.  Thus, computation of both $\nabla \svmL (\bw)$ and
$\nabla^2 \svmL (\bw)$  require the submatrix $X_{I, :}$ (as seen in
Table~\ref{table:bottleneckOperations}).

While accessing $X_{I, :}$ is naturally supported through
shared-memory random
access for multithreaded speedups (e.g., in~\cite{halloran2018matter},
which used \omp{} to speed up the \tronsvm{} quantities
in Table~\ref{table:bottleneckOperations} within
Percolator), the non-contiguous nature of this operation leads to
misaligned (i.e., not coalesced) device memory which prevents
optimal GPU use.  Furthermore, as noted in
Section~\ref{section:gpuOpt}, large memory transfers between host and
device are expensive, hindering approaches where $I$ is first computed
then a randomly accessed submatrix is created on the host and
transferred to the device.  

To overcome this challenge, we
first make use of the insight that, prior to computing $\svmL (\bw)$,
the active set $I$ may be computed
and stored entirely on the device.  With $I$ on the device, the submatrix
$X_{I,:}$ may be efficiently computed \emph{within device memory}.
Computing $I$ and $X_{I,:}$ on the device entirely decouples these
variables from host-side compute and accomplishes all simultaneous
optimizations listed in Section~\ref{section:gpuOpt}.  
The major
operations of the resulting GPU-optimized solver, called \tronsvmg{},
are listed in Table~\ref{table:tronGpuSolvers}.



{\bf Decreasing GPU-memory utilization via mixed-architecture
  speedups.}
In \tronsvmg{}, the device memory required to decouple $X_{I,:}$
from host-side compute proves prohibitive for extremely large-scale
proteomics datasets.  To remedy this, the mixed-architecture
solver, \tronsvmm{}, utilizes the GPU for heavy lifting before
using multithreading for efficient random access to $X_{I,:}$ (after $I$ is
computed) during \tron{}'s conjugate gradient procedure.  Thus,
\tronsvmm{} utilizes much less GPU
memory than \tronsvmg{}, at the expense of some speed due to fewer
operations being run on the device.  The major operations of \tronsvmm{}
are listed in Table~\ref{table:tronGpuSolvers}.
\vspace{-0.1in}
\section{Results and Discussion}\label{section:results}
\vspace{-0.1in}
All experiments were run on a dual Intel Xeon Gold 5118 compute node
with 48 computational threads, an NVIDIA Tesla V100 GPU, and 768 GB of
memory.  

{\bf Speedups for sparse features}.  The \tronlr{} GPU-optimized and
mixed-architecture solvers (described in Section~\ref{section:gpuOpt})
are referred to as \tronlrg{} and \tronlrm{}, respectively.  \tronlrg{},
\tronlrm{}, and \tronminus{} were all developed based on {\sf
  LIBLINEAR v2.30}.  Single-threaded \liblinear{} tests were run using
{\sf v2.30}.  The multithread-optimized version of \tronlr{} described
in~\cite{lee2015fast}, referred to herein as \tronlrc, was tested
using {\sf multi-core LIBLINEAR v2.30}.  All
single-threaded \tronlr{} implementations (i.e., \tronminus{},
\tronlrg, and the single-threaded optimized version of \tronlr{} in
standard \liblinear{}) were run with the same command
line parameters: \texttt{-c 4 -e 0.1 -s 0}.  Multithreaded
implementations
were run with the
additional flag \texttt{-nr i}, specifying the use of $i$ compute
threads.  As is standard practice, wallclock times were measured as
the minimum reported times over ten runs.  Training times were
measured within \liblinear{} as the time elapsed calling
\texttt{tron\_obj.tron()}.  Six datasets of varying statistics (i.e.,
number of features, instances, and nonzero elements) were downloaded
from \url{https://www.csie.ntu.edu.tw/~cjlin/libsvmtools/datasets/}
and used to benchmark the \tronlr{} solvers (statistics for
each dataset are listed in Appendix~\ref{appendix:datasets}).  

\begin{figure*}
  \centering
  \begin{subfigure}[t]{0.23\linewidth}
    \includegraphics[trim=5.3in 1.8in 0.09in
    1.3in, clip=true, width=1.\linewidth]{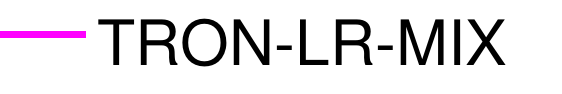}
  \end{subfigure}
  \begin{subfigure}[t]{0.23\linewidth}
    \includegraphics[trim=5.3in 1.5in 0.09in
    1.6in, clip=true, width=1.\linewidth]{rcv1Legend}
  \end{subfigure}
  \begin{subfigure}[t]{0.23\linewidth}
    \includegraphics[trim=5.3in 1.2in 0.09in
    1.9in, clip=true, width=1.\linewidth]{rcv1Legend}
  \end{subfigure}
  \begin{subfigure}[t]{0.23\linewidth}
    \includegraphics[trim=5.3in 0.9in 0.08in
    2.2in, clip=true, width=1.\linewidth]{rcv1Legend}
  \end{subfigure}
  \begin{subfigure}[t]{0.32\linewidth}
    \includegraphics[trim=0in 0in 0.33in 0in, clip=true,width=1.0\linewidth]{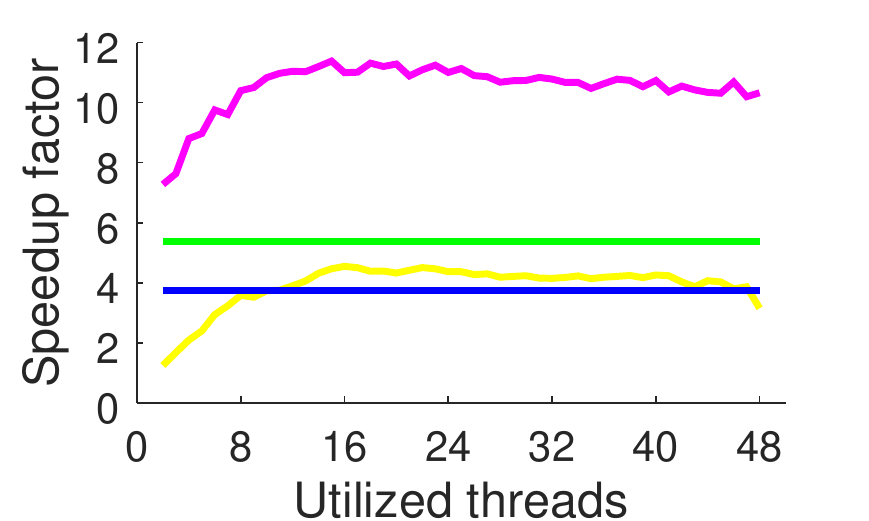}
    \caption{{\small real-sim}}
    \label{fig:real}
  \end{subfigure}
  \begin{subfigure}[t]{0.32\linewidth}
    \includegraphics[trim=0in 0in 0.33in 0in, clip=true,width=1.0\linewidth]{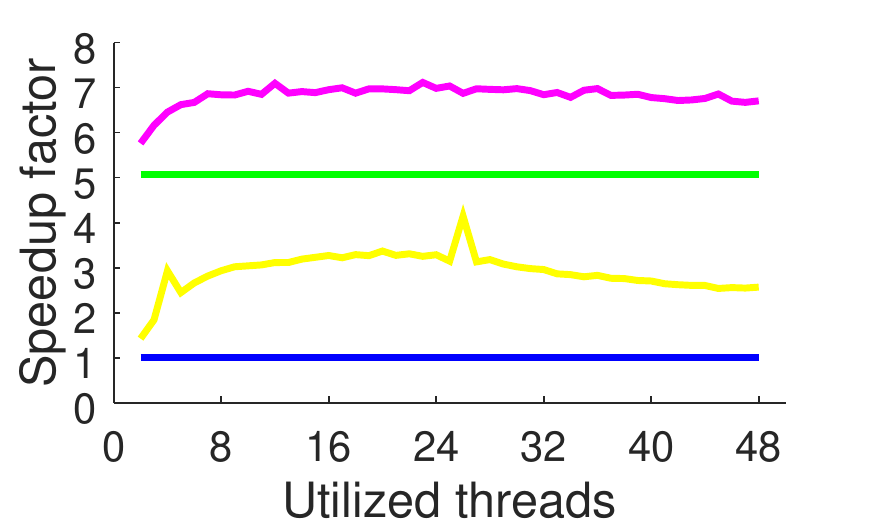}
    \caption{{\small kddb}}
    \label{fig:kddb}
  \end{subfigure}
  \begin{subfigure}[t]{0.32\linewidth}
    \includegraphics[trim=0in 0in 0.33in 0in, clip=true,width=1.0\linewidth]{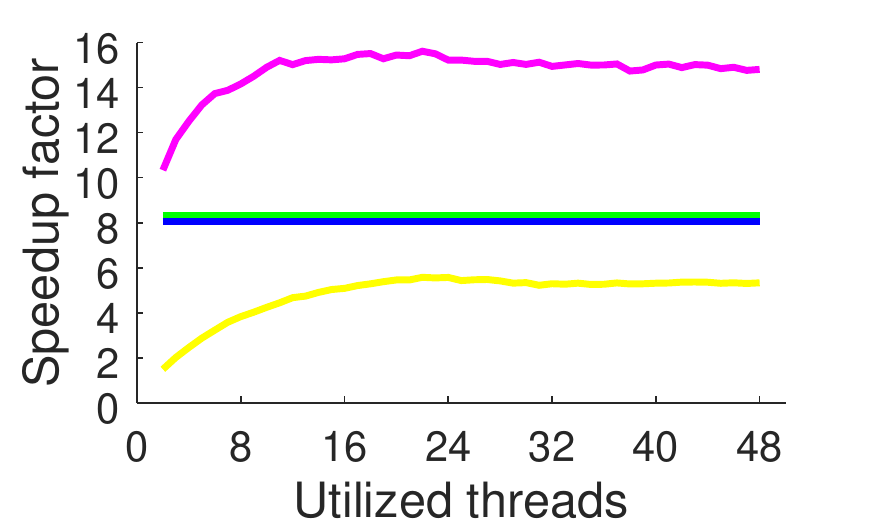}
    \caption{{\small url}}
    \label{fig:url}
  \end{subfigure}
  \begin{subfigure}[t]{0.32\linewidth}
    \includegraphics[trim=0in 0in 0.33in 0in, clip=true,width=1.0\linewidth]{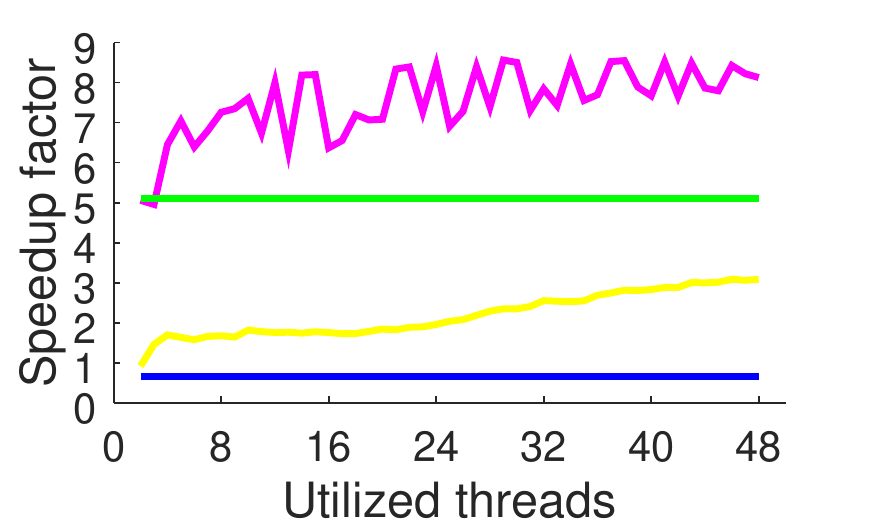}
    \caption{{\small SUSY}}
    \label{fig:susy}
  \end{subfigure}
  \begin{subfigure}[t]{0.32\linewidth}
    \includegraphics[trim=0in 0in 0.33in 0in, clip=true,width=1.0\linewidth]{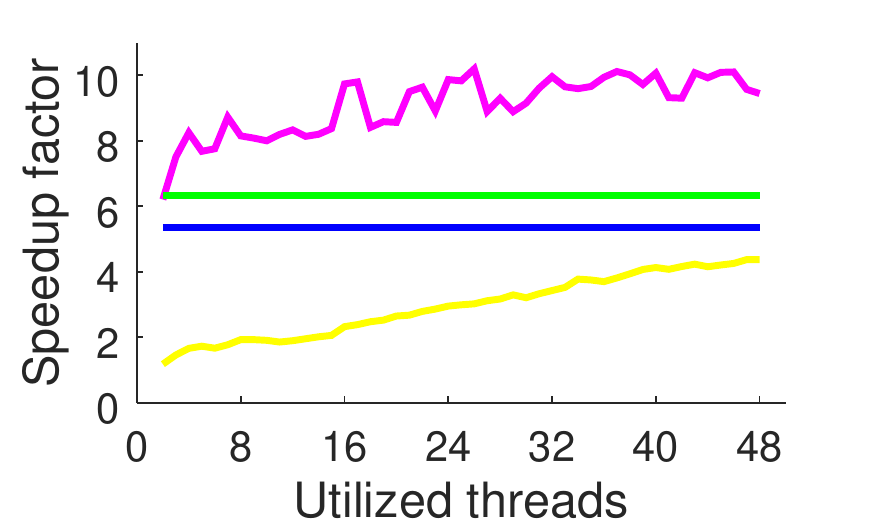}
    \caption{{\small HIGGS}}
    \label{fig:higgs}
  \end{subfigure}
  \begin{subfigure}[t]{0.32\linewidth}
    \includegraphics[trim=0in 0in 0.33in 0in, clip=true,width=1.0\linewidth]{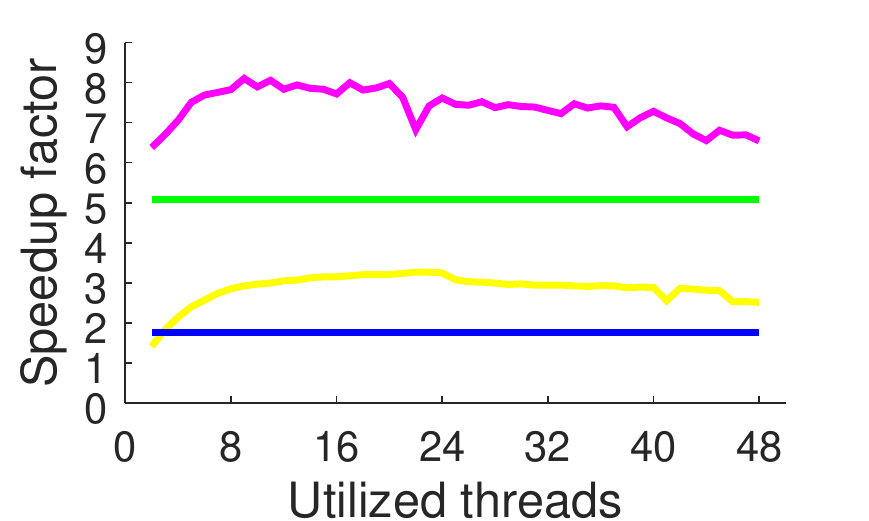}
    \caption{{\small rcv1}}
    \label{fig:rcv1}
  \end{subfigure}
  \caption{\small Factor of speedup for several optimized versions of
    \tronlr{} in \liblinear{}.  The $x$-axis
    displays the number of threads used for multithreaded methods.
    The $y$-axis denotes the multiplicative factor of training speedup
    for each method relative to the single-threaded
  version of \tronlr{} in the standard \liblinear{}.  Training times
  were measured within \liblinear{} as the time elapsed calling
  \texttt{tron\_obj.tron()}.  As is standard practice, wallclock times
  were measured as the minimum reported times over ten runs.}
  \label{fig:lrTiming}
\vspace{-0.1in}
\end{figure*}

The speedups for all methods are displayed in
Figure~\ref{fig:lrTiming}.  
\tronlrg{} significantly outperforms the multithread-optimized
\tronlrc{} and the direct GPU implementation, \tronminus{}, across all
datasets and threads.  The mixed-architecture \tronlrm{}
further improves upon \tronlrg{} performance in each dataset for all
threads used, leading to over tenfold speedups in training time
on half of the presented datasets.  We note that, due to thread
scheduling overhead, multithreaded methods experience diminished
performance for large numbers of threads in
Figures~\ref{fig:real},\ref{fig:kddb},\ref{fig:rcv1}.  However,
the presented GPU optimizations consistently provide the best speedups
when both multithreading is not used and when multithreading is
overutilized.



{\bf Speedups for dense features.}  The  \tron{} GPU
solvers described in Section~\ref{section:svmGpus}--the GPU-optimized
\tronsvmg{} and the mixed-architecture \tronsvmm{}--are compared
against the multithread-optimized versions
of \tron{} (referred to as \tronsvmc{}) and \cgls{}
from~\cite{halloran2018matter}.  The methods are tested using two
extremely large datasets.  The first dataset, referred to as the
Kim dataset, is a larger version of the benchmark dataset used
in~\cite{halloran2018matter}, consisting of 23,330,311
PSMs (i.e., proteomics data instances, described
in~\ref{section:percolator}).  The second dataset, referred
to as the Wilhelm dataset, was collected from a map of the human
proteome~\cite{wilhelm2014mass} and contains 215,282,771 PSMs.  All
multithreaded solvers were tested using 8, 16, 24, 32, 40, and 48
threads.  As in~\cite{halloran2018matter}, to effectively measure the
runtime of multithreaded methods without excess thread-scheduling
overhead, parallelization of Percolator's outermost cross-validation
was disabled.

\begin{figure}
\centering
\begin{subfigure}[t]{0.47\linewidth}
  \includegraphics[trim=60 180
  60 225,clip=true,width=1.\textwidth]{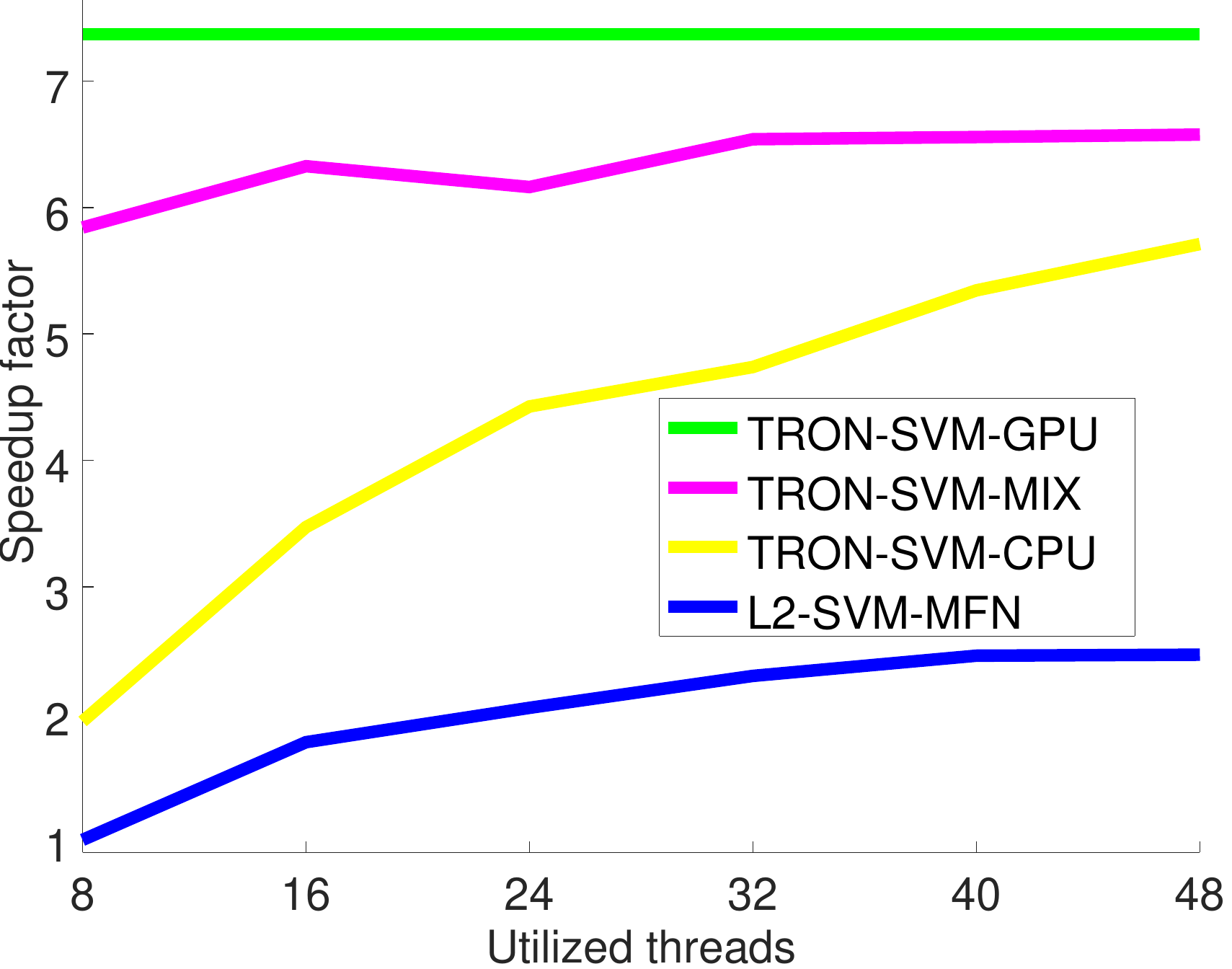}
  \caption{{\small SVM speedups for a
      large, dense proteomics dataset containing 23,330,311 PSMs.}}
  \label{figure:speedupResults}
\end{subfigure}
\;
\begin{subfigure}[t]{0.47\linewidth}
  \includegraphics[trim=60 180
  60 220,clip=true,width=1.\textwidth]{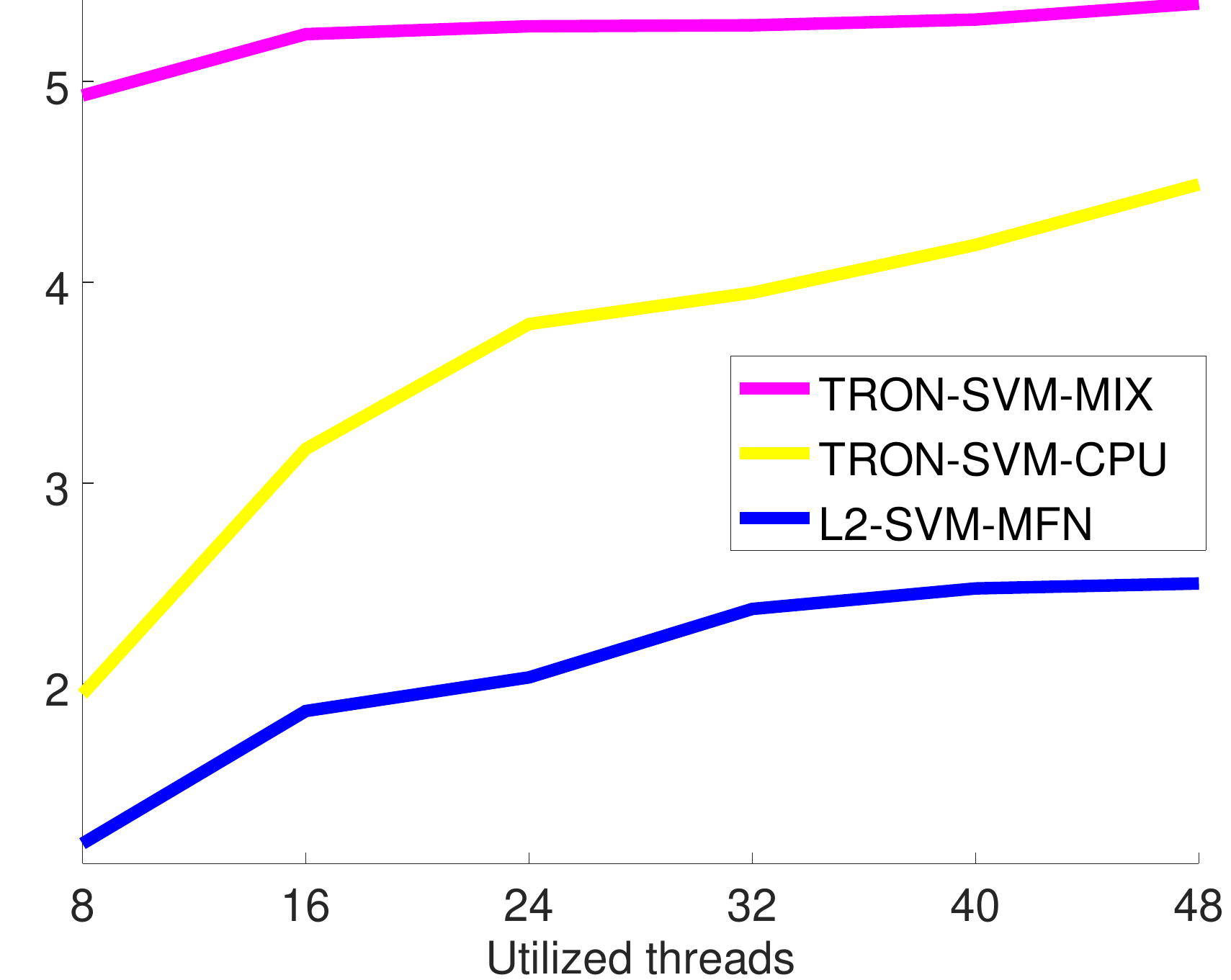}
  \caption{{\small SVM speedups for a massive dense dataset containing
      215,282,771 PSMs, too large to be analyzed by
      ``TRON-SVM-GPU.''}}
  \label{figure:svmResults2}
\end{subfigure}
\caption{\small Factor of speedup for SVM learning in Percolator for dense
  large- and massive-scale datasets.  Speedup factor is calculated as the
  original Percolator SVM learning time divided by the sped up
  learning time.  The x-axis displays the number of threads utilized
  by multithreaded methods ``L2-SVM-MFN,'' ``TRON-SVM-CPU,'' and
  ``TRON-SVM-MIX.''}
\label{figure:percolatorSpeedups}
\vspace{-0.1in}
\end{figure}

Reported runtimes are the minimum wall-clock times
measured over five runs for the Kim dataset and three runs for the
Wilhelm dataset.  The original Percolator SVM learning runtimes
(collected using {\sf Percolator v3.04.0}) were 14.4 hours and
4.4. days for the Kim and Wilhelm datasets, respectively.  Speedups
for both datasets are illustrated in
Figure~\ref{figure:percolatorSpeedups}.
For the Kim
dataset, speedup results for all discussed methods are illustrated in
Figure~\ref{figure:speedupResults}.  For the Wilhelm dataset, total
Tesla V100 memory (16 GB) is exceeded for
\tronsvmg{}.  However, the reduced memory requirements of \tronsvmm{}
allow GPU speedups for this massive dataset (illustrated in
Figure~\ref{figure:svmResults2}).


Both GPU solvers greatly accelerate Percolator SVM learning while
dominating previously proposed multithreaded speedups.  For
the Kim dataset, \tronsvmm{} and \tronsvmg{} achieve 6.6 and 7.4 fold
speedups, respectively, over Percolator's current SVM learning
engine.  
For the Wilhelm dataset, \tronsvmm{} achieves a 5.4
fold speedup while being notably efficient using few system
threads--with at most 16 threads, \tronsvmm{} improves the average
training time of \tronsvmc{} and \cgls{} by 50\% and 70\%,
respectively.
Together, these two solvers present versatile trade-offs for different
compute environments; when the dataset does not exceed the GPU memory,
\tronsvmg{} offers superior performance.  However, when onboard GPU
memory is limited, a small portion of speed may be traded
for much less memory consumption by using \tronsvmm{}.  Furthermore,
when the number of computational threads is also limited, \tronsvmm{}
offers significantly better (and more stable) performance at low
numbers of utilized threads compared to the purely multithreaded
solvers \tronsvmc{} and \cgls{}.

\vspace{-0.05in}
\section{Conclusions and Future Work}\label{section:conclusions}
\vspace{-0.05in}
In this work, we've shown that by using general GPU-optimization
principles, excellent speedups may be enjoyed by algorithms which
natively resist GPU optimization.  For the widely
used \tron{} algorithm, we've presented several GPU-optimized solvers
for both sparse and dense feature sets of \reg-regularized
primal problems.  Using a single GPU, these solvers were shown
to dominate recently proposed speedups for logistic regression (within
\liblinear{}) and SVM classification for state-of-the-art proteomics
analysis (within Percolator).  Furthermore, for sparse features,
we've shown how multithreading may compliment GPU optimizations and,
for memory-restrictive dense features, how multithreading may relieve
device-memory requirements while allowing
substantial GPU speedups.  The former optimizations
achieve over an order-of-magnitude speedup on half of the presented
datasets (and an average 9.3 fold speedup on all datasets), while the
latter optimizations decrease massive-scale biological analysis time
from 4.4 days down to just 19.7 hours.

There are significant avenues for future work.  We plan to extend
GPU-optimized \tron{} implementations to use general gradient and
Hessian-vector product information computed in auotomatic
differentiation~\cite{baydin2017automatic} packages such as
\pytorch{}~\cite{pytorchNeurips2019} and
\tensorflow{}~\cite{tensorflow2015-whitepaper}, which utilize second-order primal solvers
(such as \lbfgs{}) to optimize losses while relying on
GPU compute for optimal performance.  Furthermore, we plan to apply
the presented GPU-optimization principles to speed up other
fast machine learning solvers~\cite{hsieh2008dual,andrew2007scalable,
  yuan2012improved, keerthi2005modified, johnson2015blitz,
  johnson2018fast} which, like \tron{}, are natively designed to rely
on sequential dependencies of variables.
\section*{Broader Impact}
This paper solely focuses on speeding up machine learning software,
and thus impacts machine learning packages or applications
which use either the included software or the paper's GPU optimization
principles (to speed up an algorithm not discussed).  Benefits include faster
software, with specific applications including real-time
classification for self-driving cars~\cite{yoshioka2017real}, flagging
credit card fraud~\cite{dhankhad2018supervised}, water monitoring to
preserve ecosystems in maritime and archipelagic
countries~\cite{arridha2018classification}), etc.  Machine learning
companies/researchers/practitioners who do not use GPU resources may
be put at a disadvantage from this research, but any
advantage/disadvantage is defined solely in terms of training time
speed.

\noindent {\bf Acknowledgments}: This work was supported by the
National Center for Advancing Translational Sciences (NCATS), National
Institutes of Health, through grant UL1 TR001860 and a GPU donation
from the NVIDIA Corporation.

\bibliographystyle{plain}
\setcitestyle{numbers, open={[}, close={]}}
\bibliography{gpuOpt}
\appendix
\section{GPU speedups training a logistic regression classifier in \pytorch{}}\label{appendix:pytorch}
A binary logistic regression classifier was implemented in \pytorch{} ({\sf
  v1.4.0} ) and trained over the \texttt{rcv1} dataset to
illustrate the speed ups possible using a GPU (Nvidia Tesla V100)
versus only  multithreading (24 CPU threads using an Intel Xeon Gold
5118).  Speedups were tested for both batch gradient descent (with a
0.001 learning rate) and
\lbfgs{}.  The \texttt{rcv1} dataset was downloaded from
\url{https://www.csie.ntu.edu.tw/~cjlin/libsvmtools/datasets/binary/rcv1_train.binary.bz2}.
Gradient descent converged after 3,000 iterations and \lbfgs{}
converged after 100 iterations.  For reference, a logistic regression
classifier was trained using single-threaded \tron{} (as implemented in \sklearn{}
{\sf v0.20.4}).  All code is available in
  \texttt{pyTorchLogisticRegression\_rcv1.py}.
\begin{table}[htbp!]
\centering
\begin{tabular}{l|c|c|c}
Solver & CPU training time (s) & GPU training time (s) & GPU Speedup\\\hline
Gradient descent & 395.58 & 27.05 & 14.63\\\hline
\lbfgs{} & 40.56 & 3.1 & 13.08 \\\hline
\tron{} (\sklearn{}) & 0.29 & -- & -- \\\hline
\end{tabular}
\vspace{0.1in}
\caption{Logistic regression training times, measured in seconds,
  for the \texttt{rcv1} dataset.  Gradient descent and \lbfgs{}
  solvers are implemented in \pytorch{}, and single-threaded \tron{}
  is implemented in \sklearn{}.}
\label{table:pytorchTrainingTimes}
\end{table}

\section{Derivation of \tron{} Hessian-vector products}\label{section:derivation}
Consider feature vectors $\bx_i \in \RR^n, i = 1, \dots, l$ and label vector
$\by \in \{-1, 1\}^l$, and  let $X = [\bx_1 \dots 
  \bx_l]^T$ be the feature matrix.  For vectors, index-set subscripts
  denote subvectors and for matrices,
pairs of index-set subscripts denote submatrices. Let $\indicator$ denote the indicator function.

The general \reg-regularized objective, which we wish to minimize w.r.t. $\bw$, is
\begin{align}
f(\bw) =& \frac{1}{2}\bw^T\bw + C \sum_{i = 1}^l \loss(\bw;\bx_i, y_i),\label{eq:loss2}
\end{align}
where $\frac{1}{2}\bw^T\bw$ is the regularization term, $C > 0$ is a
regularization hyperparameter, and $\loss(\bw;\bx_i, y_i)$ is a loss
function.  When $\loss(\bw;\bx_i, y_i) = \log ( 1 +
\exp{ (-y_i \bw^T \bx_i) }$, commonly referred to as the logistic
loss, minimizing Equation~\ref{eq:loss2} corresponds to learning a
classifier using logistic regression.  Similarly, minimizing
Equation~\ref{eq:loss2} when $\loss(\bw;\bx_i, y_i) = (\max(0,1 - y_i
\bw^T \bx_i))^2$, commonly referred to as the \lsvm{} or quadratic SVM
loss, corresponds to learning a linear SVM classifier.  The logistic
loss results in an objective function that is twice
differentiable and the \lsvm{} loss yields a differentiable
objective (unlike the hinge loss) with a generalized
Hessian~\cite{keerthi2005modified}).  We denote Equation~\ref{eq:loss2}
under the logistic loss as  $\lrL$ and, under the \reg-SVM loss, as
$\svmL$.

\tron{} is detailed in Algorithm~\ref{algorithm:tron}.  At each
iteration, given the current parameters $\bw$ and trust region
interval $\Delta$, TRON considers the following quadratic
approximation to $f(\bw + \bd) - f(\bw)$,
\begin{align}
q(\bd) = \nabla
f(\bw)^T\bd + \frac{1}{2}\bd^T \nabla^2
f(\bw)\bd.\label{eq:quadraticApprox2}
\end{align}
A truncated Newton step, confined in the trust region, is then found
by solving
\begin{align}\label{eq:minQuadraticApprox}
\min_{\bd}q(\bd) \quad & \mbox{ s.t. } \lVert \bd \rVert_2 \leq \Delta.
\end{align}
If $q(\bd)$ is close to $f(\bw + \bd)
- f(\bw)$, $\bw$ is updated to $\bw + \bd$ and the trust region
interval is increased for the subsequent iteration.  Otherwise, $\bw$
remains unchanged and the trust region interval is shrunk.  

\begin{algorithm}
\caption{The \tron{} algorithm}\label{algorithm:tron2}
\begin{algorithmic}[1]\small
\State Given $w$, $\Delta$, and $\sigma_0$
\State Calculate $f(\bw)$ \Comment{Critically depends on $\bz =
  X^T\bw$}
\While{Not converged}
\State Find $\bd =  \argmin_{\bv}q(\bv)\, \mbox{ s.t. }
\lVert \bv \rVert_2 \leq \Delta.$ \LONGCOMMENT{Critically depends on
  $\nabla f(\bw)$, $\nabla^2 f(\bw) \bv$}
\State Calculate $f(\bw + \bd), \sigma = \frac{f(\bw + \bd) -
  f(\bw)}{q(\bd)}$ \Comment{Critically depends on $\bz =
  X^T(\bw + \bd)$}
\If{$\sigma > \sigma_0$}
\State $\bw \leftarrow \bw + \bd$, increase trust region $\Delta$.
\Else \quad 
\State Shrink $\Delta$.
\EndIf
\EndWhile
\end{algorithmic}
\end{algorithm}

Note that the function evaluation $f(\bw)$ must be computed for each new
iteration, as well as the gradient and the Hessian for
Equation~\ref{eq:quadraticApprox2}.  However,
Equation~\ref{eq:quadraticApprox2} involves only a
Hessian-vector product, computation of which circumvents loading the
entire Hessian into memory.
For the logistic loss, we have  
\begin{align}
\nabla
\lrL (\bw) = \bw + C
\sum_{i=1}^l (h(y_i \bw^T \bx_i) - 1) y_i \bx_i, 
\end{align}
where $h(y_i \bw^T\bx_i) = (1 + e^{-y_i \bw^T\bx_i})^{-1}$.  For the
\lsvm{} loss, we have 
\begin{align}
\nabla \svmL (\bw) = \bw +  2C\hat{X}^T\hat{\bz}= \bw +
2CX_{I, :}^T(X_{I, :}\bw - \by_{I}),
\end{align}
 where $I \equiv \{ i | 1 - y_i
\bw^T \bx_i > 0 \}$ is an index set and and the operator $:$ denotes
all elements along the corresponding dimension (i.e., all columns in
this case).  Thus, $X_{I,:}$ is the submatrix of all $X$ rows the
indices of which are in $I$.

Equation~\ref{eq:minQuadraticApprox} involves only a single
Hessian-vector product, the structure of which is exploited to
avoid loading the entire Hessian into memory.
For the logistic loss, we have
\begin{align}
\nabla^2 \lrL (\bw)  =& \Id + C
X^T\lrD X,
\end{align}
where $\lrD$ is a diagonal matrix with elements $\lrD_{i,i} = h(y_i \bw^T
\bx_i)(1 - h(y_i \bw^T \bx_i))$.  Thus, for a
vector $\bv$, the Hessian-vector product is efficiently computed as
$\nabla^2 \lrL (\bw) \bv = \bv + C
X^T(\lrD (X\bv))$.
For the \lsvm{} loss, we have 
\begin{align}
\nabla^2 \svmL (\bw)  = &\Id + 2 C
X^T \svmD X= \Id + 2 C
X^T_{I,:}X_{I,:},
\end{align}
where $\svmD$ is a diagonal matrix with elements $\svmD_{i,i} = \indicator_{i \in
  I}$.  
The Hessian-vector product is thus efficiently computed as
$\nabla^2 \svmL (\bw) \bv = \bv + 2 C
X_{I,:}^T(X_{I,:}\bv)$.

\section{Concealing large-memory transfer latency between the host
  and device}\label{section:concealing}
To optimally conceal device-to-host transfer latency while maximizing
host and device parallelism, it is necessary to:
\begin{enumerate}[label=(\alph*),topsep=-1pt,itemsep=-0.1ex]
\item Add all dependent device-functions involving the data to be sent
  to an asynchronous device stream, $s$,
\item add the transfer of the data from device-to-host to $s$,
\item run independent host and/or device operations,
\item synchronize $s$ just prior to running a dependent operation on
  the host.
\item Note that if the dependent data needed from the device on the
  host is a scalar, it may be returned without latency.
\end{enumerate}

The other direction is slightly different.  To optimally conceal host-to-device transfer latency while maximizing
host and device parallelism, it is necessary to:
\begin{enumerate}[label=(\alph*),topsep=-1pt,itemsep=-0.1ex]
\item launch the transfer on a device stream as soon as the data is
  available,
\item add all dependent device-functions involving the data being sent
  to the device stream.
\end{enumerate}
It is easy to see that algorithms with many sequential dependencies
are at odds with these principles (they reveal transfer latency while
minimizing host/device parallelism).

\section{Optimization of \tronlr{} Hessian-vector products for GPUs}\label{section:hessianVectorProducts}
We complete the total GPU-optimization of \tronlr{} by considering the
remaining bottleneck, the Hessian-vector product $\nabla^2 \lrL (\bw)
\bv = \Id + C X^T( \lrD (X \bv))$.  As with the previous
optimizations, device variables are maximally decoupled from host-side
dependencies, while using device-side functions which
allow peak performance.  In particular, we compute the diagonal
matrix $\lrD$ in the same custom \cuda{} kernel used to compute
$\hat{\bz}$ (where $\lrD_{i,i} = h(y_i \bz_i)(1 - h(y_i \bz_i))$.  $\lrD$
is also used in later host computations (for
preconditioning~\cite{hsia2018preconditioned}), so $\lrD$ is
immediately transferred from device to host on an asynchronous device
stream (the stream is synchronized just prior to host-variable use).  

The candidate Newton step $\bv$ (which is only of dimension $n$)
is transferred from device to host on an asynchronous stream, and the
following decompositions of $\nabla^2 \lrL (\bw)
\bv$ are added to this same stream: $\ba_0 = X \bv, \ba_1 = \lrD
\ba_0, \ba_2 = C X^T \ba_1$.  $\ba_0$ and $\ba_2$ are computed using
\cusparse{}, while $\ba_1$ is computed using a custom kernel
for element-wise multiplication along $\lrD$'s diagonal.  $\nabla^2
\lrL (\bw) \bv$ is then transferred from host to device.  However, the
rest of the conjugate procedure is sequentially dependent on the
dot-product $\bv^T \nabla^2 \lrL (\bw) \bv$.  In order to relieve this
dependence while the $\nabla^2 \lrL (\bw) \bv$ transfers from device
to host, $\bv^T \nabla^2 \lrL (\bw) \bv$ is computed on the device and
the resulting scalar is available immediately to the host.

\section{Summary of major \tronlrg{} operations}\label{section:summary}
The following summarizes the major operations of the GPU-optimized \tron{} logistic regression
solver, \tronlrg{}, as described in the main paper and herein.  For each set of
operations, the original lines from Algorithm~\ref{algorithm:tron2} being
optimized are listed in red.
\begin{itemize}[leftmargin=5.5mm]
\item $\bz =X\bw$ is calculated and stored on the device
  \algred{(lines 2 and 5)}.
\item The vectors $\balpha$, $\bzhat$ and diagonal matrix $D$ are calculated on the device, such
that $\balpha = \log ( 1 / h(y_i \bz_i))$, $\bzhat_i = (h(y_i
\bz_i) - 1) y_i$, and $\lrD_{i,i} = h(y_i \bz_i)(1 - h(y_i \bz_i))$,
where $h(y_i\bz_i) = (1 + e^{-y_i\bz_i})^{-1}$ \algred{(lines 2 and 5)}.  $D$ is asynchronously
transferred back to the host for future preconditioning computations.
\item With $\balpha$ in device memory, the objective $\lrL (\bw) =  \frac{1}{2}\bw^T\bw +
C \sum_{i = 1}^l\log ( 1 + \exp{ (-y_i \bz_i) )} = \frac{1}{2}\bw^T\bw
+ C \sum_{i = 1}^l \balpha$ is computed \algred{(lines 2 and 5)}.
\item With $\bzhat$ in device memory, the gradient $\nabla \lrL(\bw) = \bw +
X^T \bzhat$ is computed and transferred asynchronously back to the
host \algred{(line 7)}.
\item While all the above device-side quantities are being computed, the host runs
independent, sequential operations concurrently, synchronizing the
transfer streams for $D$ and $\nabla \lrL(\bw)$ just prior to
host-side use \algred{(lines 7 and 4, respectively)}.
\item The Hessian-product is computed on the device as $\nabla^2 \lrL (\bw)
\bv = \bv + C X^T(\lrD (X\bv))$.  Subsequently, the vector-Hessian-vector
product $\bv^T \nabla^2 \lrL (\bw)\bv$ is computed on the device and the
resulting scalar is immediately available to the host \algred{(line 4)}.
\end{itemize}

\section{Benchmark Dataset Statistics}\label{appendix:datasets}
\begin{table}[htbp!]
\centering
\begin{tabular}{l|rrr}
Dataset & \#instances & \#features & \#nonzeros\\\hline
\texttt{rcv1} & 20,242 & 47,236 & 1,498,952\\\hline
\texttt{SUSY} & 5,000,000 &  18 & 88,938,127 \\\hline
\texttt{HIGGS} & 11,000,000 & 28& 283,685,620\\\hline
\texttt{KDD2010-b} & 19,264,097 & 29,890,095 & 566,345,888\\\hline
\texttt{url} & 2,396,130 & 3,231,961& 277,058,644\\\hline
\texttt{real-sim} & 72,309 & 20,958 & 3,709,083\\\hline
\hline
\texttt{Kim} & 23,330,311 & 18 & 419,945,598\\\hline
\texttt{Wilhelm} & 215,282,771 & 18 & 3,875,089,878\\\hline
\end{tabular}
\vspace{0.1in}
\caption{Sparse and dense benchmark dataset statistics for \tronlr{}
  and \tronsvm{}, respectively.}
\label{table:datasets}
\end{table}
\end{document}